# Sufficient conditions for convergence of Loopy Belief Propagation

**Joris M. Mooij**
Dept. of Biophysics, Inst. for Neuroscience
Radboud University Nijmegen
6525 EZ Nijmegen, The Netherlands
j.mooij@science.ru.nl

**Hilbert J. Kappen**
Dept. of Biophysics, Inst. for Neuroscience
Radboud University Nijmegen
6525 EZ Nijmegen, The Netherlands
b.kappen@science.ru.nl

## Abstract

We derive novel sufficient conditions for convergence of Loopy Belief Propagation (also known as the Sum-Product algorithm) to a unique fixed point. Our results improve upon previously known conditions. For binary variables with (anti-)ferromagnetic interactions, our conditions seem to be sharp.

## 1 Introduction

Loopy Belief Propagation (LBP), also known as the Sum-Product Algorithm or simply Belief Propagation, is an important algorithm for approximate inference on graphical models. Spectacular applications can be found e.g. in coding theory (iterative decoding algorithms for Turbo Codes and Low Density Parity Check Codes) and in combinatorial optimization (the Survey Propagation algorithm used in satisfiability problems such as 3-SAT and graph colouring). LBP has been generalized in many ways (e.g. double-loop algorithms, GBP, EP/EC).

However, there are two major problems in applying LBP to concrete problems: (i) if LBP converges, it is not clear whether the results are a good approximation of the exact marginals; (ii) LBP does not always converge, and in these cases one gets no approximations at all. For LBP, the two issues might be interrelated: the "folklore" is that failure of LBP to converge indicates low quality of the Bethe approximation on which it is based. This would mean that if one has to "force" LBP to converge (e.g. by using damping or double-loop approaches), one can expect the results to be of low quality.

Although LBP is an old algorithm that has been reinvented many times in different fields, a solid theoretical understanding of the two aforementioned issues is still lacking, except for the special case of a network containing a single loop [4]. Considerable progress has been made recently regarding the question under what conditions LBP converges [1, 2] and under what conditions the fixed point is unique [3].

In this work, we study the question of convergence of LBP and derive new sufficient conditions for LBP to converge to a unique fixed point, which are directly formulated in terms of arbitrary factor graphs. We present our results in a sketchy way and refer to [5] for the details.

## 2 Background

### 2.1 Graphical models and factor graphs

Consider $N$ discrete random variables $x_i \in \mathcal{X}_i$, for $i \in V := \{1, 2, \ldots, N\}$; we write $x = (x_1, \ldots, x_N) \in \mathcal{X} := \prod_{i \in V} \mathcal{X}_i$. We are interested in the following probability measure on $\mathcal{X}$:[1]

$$P(x_1, \ldots, x_N) := \frac{1}{Z} \prod_{I \in \mathcal{I}} \psi^I(x_I) \qquad (1)$$

which factorizes in *factors* or *potentials* $\psi^I$. The factors are indexed by subsets of $V$, i.e. $\mathcal{I} \subseteq \mathcal{P}(V)$. If $I \in \mathcal{I}$ is the subset $I = \{i_1, \ldots, i_m\} \subseteq V$, we write $x_I := (x_{i_1}, \ldots, x_{i_m}) \in \prod_{i \in I} \mathcal{X}_i$. Each factor $\psi^I$ is a positive function $\psi^I : \prod_{i \in I} \mathcal{X}_i \to (0, \infty)$. $Z$ is a normalizing constant ensuring that $\sum_{x \in \mathcal{X}} P(x) = 1$. We will use uppercase letters for indices of factors $(I, J, K, \ldots \in \mathcal{I})$ and lowercase letters for indices of variables $(i, j, k, \ldots \in V)$. The *factor graph* that corresponds to the probability distribution (1) is a bipartite graph with vertex set $V \cup \mathcal{I}$. In the factor graph, each *variable node* $i \in V$ is connected with all the factors $I \in \mathcal{I}$ that contain the variable, i.e. the neighbours of $i$ are the factor nodes $N_i := \{I \in \mathcal{I} : i \in I\}$. Similarly, each *factor node* $I \in \mathcal{I}$ is connected with all the vari-able nodes $i \in V$ that it contains and we will simply denote the neighbours of $I$ by $I = \{i \in V : i \in I\}$.

---

[1] The class of probability measures described by (1) contains Markov Random Fields as well as Bayesian Networks.

## 2.2 Loopy Belief Propagation

Loopy Belief Propagation is an algorithm that calculates approximations to the marginals $\{P(x_I)\}_{I \in \mathcal{I}}$ and $\{P(x_i)\}_{i \in V}$ of the probability measure (1). The calculation is done by message-passing on the factor graph: each node passes messages to its neighbours. One usually discriminates between two types of messages: messages $\mu^{I \to i}(x_i)$ from factors to variables and messages $\mu^{i \to I}(x_i)$ from variables to factors (where $i \in I \in \mathcal{I}$). Both messages are positive functions on $\mathcal{X}_i$, or, equivalently, vectors in $\mathbb{R}^{\mathcal{X}_i}$ (with positive components). The messages that are sent by a node depend on the incoming messages; the new messages, designated by $\tilde{\mu}$, are given in terms of the incoming messages by the following *LBP update rules*[2]

$$\tilde{\mu}^{j \to I}(x_j) \propto \prod_{J \in N_j \setminus I} \mu^{J \to j}(x_j) \quad (2)$$

$$\tilde{\mu}^{I \to i}(x_i) \propto \sum_{x_{I \setminus i}} \psi^I(x_I) \prod_{j \in I \setminus i} \mu^{j \to I}(x_j). \quad (3)$$

Usually, one normalizes the messages in the $\ell_1$-sense (i.e. such that $\sum_{x_i \in \mathcal{X}_i} \mu(x_i) = 1$). When all messages have converged to some fixed point $\mu_\infty$, one calculates the approximate marginals or *beliefs*

$$b_i(x_i) = C^i \prod_{I \in N_i} \mu_\infty^{I \to i}(x_i) \approx P(x_i)$$

$$b_I(x_I) = C^I \psi^I(x_I) \prod_{i \in I} \mu_\infty^{i \to I}(x_i) \approx P(x_I),$$

where the $C^i$'s and $C^I$'s are normalization constants ensuring that the approximate marginals are normalized in $\ell_1$-sense. Note that the beliefs are invariant under rescaling of the messages, which shows that the precise way of normalization in (2) and (3) is irrelevant. For numerical stability however, some way of normalization (not necessarily in $\ell_1$-sense) is desired to ensure that the messages stay in some compact domain.

In the following, we will formulate everything in terms of the messages $\mu^{I \to i}(x_i)$ from factors to variables; the update equations are then obtained by substituting (2) in (3):

$$\tilde{\mu}^{I \to i}(x_i) = C^{I \to i} \sum_{x_{I \setminus i}} \psi^I(x_I) \prod_{j \in I \setminus i} \prod_{J \in N_j \setminus I} \mu^{J \to j}(x_j). \quad (4)$$

---

[2]We abuse notation slightly by writing $X \setminus x$ instead of $X \setminus \{x\}$ for sets $X$.

with $C^{I \to i}$ such that $\sum_{x_i \in \mathcal{X}_i} \tilde{\mu}^{I \to i}(x_i) = 1$. We consider here LBP with a *parallel* update scheme, which means that all message updates (4) are done in parallel.

## 3 Special case: binary variables with pairwise interactions

In this section we investigate the simple special case of binary variables (i.e. $|\mathcal{X}_i| = 2$ for all $i \in V$), and in addition we assume that all potentials consist of at most two variables. Although this is a special case of the more general theory to be presented later on, we start with this simple case because it illustrates most of the underlying ideas without getting involved with the additional technicalities of the general case.

Without loss of generality we assume that $\mathcal{X}_i = \{-1, +1\}$ for all $i \in V$. We take the factor index set as $\mathcal{I} = \mathcal{I}_1 \cup \mathcal{I}_2$ with $\mathcal{I}_1 = V$ (the "local evidence") and $\mathcal{I}_2 \subseteq \{\{i, j\} : i, j \in V, i \neq j\}$ (the "pair-potentials"). The probability measure (1) can then be written as

$$P(x) = \frac{1}{Z} \exp \left( \sum_{\{i,j\} \in \mathcal{I}_2} J_{ij} x_i x_j + \sum_{i \in \mathcal{I}_1} \theta_i x_i \right)$$

for some choice of the parameters $J_{ij}$ and $\theta_i$.

Note that the messages sent from single-node factors $\mathcal{I}_1$ to variables are constant. Thus the question whether messages converge can be decided by only studying the messages sent from pair-potentials $\mathcal{I}_2$ to variables. In this special case, it is advantageous to use the following "natural" parameterization of the messages

$$\tanh \nu^{i \to j} := \mu^{\{i,j\} \to j}(x_j = 1) - \mu^{\{i,j\} \to j}(x_j = -1), \quad (5)$$

where $\nu^{i \to j} \in \mathbb{R}$ is now interpreted as a message sent from variable $i$ to variable $j$ (instead of a message sent from the factor $\{i, j\}$ to variable $j$). The LBP update equations (4) become particularly simple in this parameterization:

$$\tanh \tilde{\nu}^{j \to i} = \tanh(J_{ij}) \tanh \left( \theta_j + \sum_{k \in N_j \setminus i} \nu^{k \to j} \right) \quad (6)$$

where we wrote $N_j := \{k \in V : \{j, k\} \in \mathcal{I}_2\}$ for the variables that interact with $j$ via a pair-potential.

Defining the set of ordered pairs $D := \{i \to j : \{i, j\} \in \mathcal{I}_2\}$, we see that the parallel LBP update is a mapping $f : \mathbb{R}^D \to \mathbb{R}^D$; the component $j \to i$ of $f(\nu) = \tilde{\nu}$ is specified in terms of the components of $\nu$ in (6). Our goal is now to derive sufficient conditions under which the mapping $f$ is a contraction. For this we need some elementary but powerful mathematical theorems.

## 3.1 Normed spaces, contractions and bounds

Let $d$ be a metric on a set $X$. A mapping $f : X \to X$ is called a *contraction* (with respect to $d$) if there exists $0 \le K < 1$ such that $d(f(x), f(y)) \le K d(x,y)$ for all $x, y \in X$. If the metric space $(X, d)$ is *complete*, we can apply the

**Theorem 3.1 (Contracting Mapping Principle)**
*Let $f : X \to X$ be a contraction of a complete metric space $(X, d)$. Then $f$ has a unique fixed point $x_\infty \in X$ and for any $x_0 \in X$, the sequence $n \mapsto x_n := f(x_{n-1})$ converges to $x_\infty$. The rate of convergence is at least linear: $d(f(x), x_\infty) \le K d(x, x_\infty)$ for $x \in X$.* □

This theorem is due to Banach; the proof can be found in many advanced textbooks on analysis.

In the following, let $V$ be a finite dimensional vector space and $\|\cdot\|$ a norm on $V$. The norm induces a *matrix norm* (also called *operator norm*) on linear mappings $A : V \to V$, defined by

$$\|A\| := \sup_{\substack{v \in V, \\ \|v\| \le 1}} \|Av\|.$$

Let $f : V \to V$ be a differentiable mapping. For $x, y \in V$, we define the segment joining $x$ and $y$ by $[x, y] := \{\lambda x + (1-\lambda)y : \lambda \in [0,1]\}$. We have the following consequence of the Mean Value Theorem:

**Lemma 3.1** *Let $x, y \in V$. Then:*

$$\|f(y) - f(x)\| \le \|y - x\| \cdot \sup_{z \in [x,y]} \|f'(z)\|$$

*where $\|f'(z)\|$ is the induced matrix norm of the derivative ("Jacobian") of $f$ at $z$.*

**Proof.** See [6, Thm. 8.5.4]. □

The norm $\|\cdot\|$ on $V$ induces a metric $d(x,y) := \|y - x\|$ on $V$. The metric space $(V, d)$ is complete, since $V$ is finite-dimensional. If $f : V \to V$ is a contraction with respect to the induced metric $d$, we say that $f$ is a $\|\cdot\|$-contraction. Combining Theorem 3.1 and Lemma 3.1 yields our basic tool:

**Lemma 3.2** *If $\sup_{v \in V} \|f'(v)\| < 1$, then $f$ is a $\|\cdot\|$-contraction and the consequences of Theorem 3.1 hold.* □

## 3.2 Sufficient conditions for LBP to be a contraction

We apply Lemma 3.2 to the case at hand: the parallel LBP update mapping $f : \mathbb{R}^D \to \mathbb{R}^D$, written out in components in (6). The derivative of $f$ is easily calculated from (6) and is given by[3]

$$\left(f'(\nu)\right)_{j \to i, k \to l} = \frac{\partial \tilde{\nu}^{j \to i}}{\partial \nu^{k \to l}} = G_{j \to i}(\nu) A_{j \to i, k \to l}$$

where

$$G_{j \to i}(\nu) := \frac{1 - \tanh^2(\theta_j + \sum_{t \in N_j \setminus i} \nu^{t \to j})}{1 - \tanh^2(\tilde{\nu}^{j \to i}(\nu))} \operatorname{sgn} J_{ij}$$

$$A_{j \to i, k \to l} := \tanh|J_{ij}| \, \delta_{j,l} \mathbf{1}_{N_j \setminus i}(k). \tag{7}$$

Note that we have absorbed all $\nu$-dependence in the factor $G_{j \to i}$; the factor $A_{j \to i, k \to l}$ is independent of $\nu$ and captures the structure of the graphical model. Note further that $\sup_{\nu \in V} |G_{j \to i}(\nu)| = 1$, implying that

$$\left|\frac{\partial \tilde{\nu}^{j \to i}}{\partial \nu^{k \to l}}\right| \le A_{j \to i, k \to l} \tag{8}$$

everywhere on $V$.

We still have the freedom to choose the norm. As an example, we take the $\ell_1$-norm, which is defined on $\mathbb{R}^n$ by $\|x\|_1 := \sum_{i=1}^n |x_i|$. The corresponding matrix or operator norm, for a linear $A : \mathbb{R}^n \to \mathbb{R}^m$, is given by $\|A\|_1 := \max_{j \in \{1, \ldots, n\}} \sum_{i=1}^m |A_{ij}|$. Using the $\ell_1$-norm on $\mathbb{R}^D$, we obtain:

**Corollary 3.1** *If*

$$\max_{l \in V} \max_{k \in N_l} \sum_{i \in N_l \setminus k} \tanh|J_{il}| < 1, \tag{9}$$

*LBP is a $\ell_1$-contraction and converges to a unique fixed point, irrespective of the initial messages.*

**Proof.** Using (8) and (7):

$$\|f'(\nu)\|_1 = \max_{k \to l} \sum_{j \to i} \left|\frac{\partial \tilde{\nu}^{j \to i}}{\partial \nu^{k \to l}}\right|$$

$$\le \max_{k \to l} \sum_{j \to i} \tanh|J_{ij}| \, \delta_{jl} \mathbf{1}_{N_j \setminus i}(k)$$

$$= \max_{l \in V} \max_{k \in N_l} \sum_{i \in N_l \setminus k} \tanh|J_{il}|$$

and now simply apply Lemma 3.2. □

## 3.3 Beyond norms: the spectral radius

Another norm that gives an analytical result is e.g. the $\ell_\infty$ norm; however, it turns out that this norm gives a weaker sufficient condition than the $\ell_1$ norm does. Instead of pursuing a search for the optimal norm, we can derive a criterion for convergence based on the spectral radius of the matrix (7). The key idea is to look at several iterations of LBP at once. This will yield a significantly stronger condition for convergence of LBP to a unique fixed point. However, this comes at a computational cost: the condition involves calculating (or bounding) the largest eigenvalue of a (possibly quite large, but sparse) matrix.

---

[3] For a set $X$, we define the indicator function $\mathbf{1}_X$ of $X$ by $\mathbf{1}_X(x) = 1$ if $x \in X$ and $\mathbf{1}_X(x) = 0$ if $x \notin X$.

For a square matrix $A$, we denote by $\sigma(A)$ its *spectrum*, i.e. the set of eigenvalues of $A$. By $\rho(A)$ we denote its *spectral radius*, which is defined as $\rho(A) := \sup |\sigma(A)|$, i.e. the largest magnitude of the eigenvalues of $A$.

**Theorem 3.2** *Let $f : \mathbb{R}^m \to \mathbb{R}^m$ be differentiable and suppose that $f'(x) = G(x)A$, where $G$ is diagonal with bounded entries $|G_{ii}(x)| \leq 1$ and $0 \leq A_{ij}$. If $\rho(A) < 1$ then for any $x_0 \in \mathbb{R}^m$, the sequence $x_0, f(x_0), f^2(x_0), \ldots$ obtained by iterating $f$ converges to a fixed point $x_\infty$, which does not depend on $x_0$.*

**Proof.** For a matrix $B$, we will denote by $|B|$ the matrix with entries $|B|_{ij} = |B_{ij}|$. For two matrices $B, C$ we will write $B \leq C$ if $B_{ij} \leq C_{ij}$ for all entries $(i, j)$. Note that if $B \leq C$, then $\|B\|_1 \leq \|C\|_1$. Also note that $|BC| \leq |B| |C|$. Finally, if $0 \leq A$ and $B \leq C$, then $AB \leq AC$ and $BA \leq CA$.

Using these observations and the chain rule, we have for any $n = 1, 2, \ldots$ and any $x \in \mathbb{R}^m$:

$$|(f^n)'(x)| = \left| \prod_{i=1}^n f'(f^{i-1}(x)) \right|$$
$$\leq \prod_{i=1}^n \Big( |G(f^{i-1}(x))| \, A \Big) \leq A^n.$$

Hence $\|(f^n)'(x)\|_1 \leq \|A^n\|_1$ for all $n = 1, 2, \ldots$ and all $x \in \mathbb{R}^m$. By the Gelfand Spectral Radius Theorem (which holds in fact for *any* matrix norm), $\lim_{n \to \infty} \|A^n\|_1^{1/n} = \rho(A)$. Now if $\rho(A) < 1$, it follows that for some $N$, $\|A^N\|_1^{1/N} < 1$. Hence $\sup_{x \in \mathbb{R}^m} \|(f^N)'(x)\|_1 < 1$. Applying Lemma 3.2, we conclude that $f^N$ is a $\ell_1$-contraction and thus has a unique fixed point $x_\infty$.

Take any $x_0 \in \mathbb{R}^m$. Consider the $N$ sequences obtained by iterating $f^N$, starting respectively in $x_0$, $f(x_0)$, ..., $f^{N-1}(x_0)$. Each sequence converges to $x_\infty$ since $f^N$ is a $\ell_1$-contraction with fixed point $x_\infty$. But then the sequence $x_0, f(x_0), f^2(x_0), \ldots$ must converge to $x_\infty$. $\square$

This immediately yields:

**Corollary 3.2** *For binary variables with pairwise interactions, LBP converges to a unique fixed point, irrespective of the initial messages, if the spectral radius of the $|D| \times |D|$-matrix*

$$A_{j \to i, k \to l} := \tanh |J_{ij}| \, \delta_{j,l} \mathbf{1}_{N_j \setminus i}(k)$$

*is strictly smaller than 1.* $\square$

One might think that there is a shorter route to this result: it seems quite plausible intuitively that in general, for a continuously differentiable $f : \mathbb{R}^m \to \mathbb{R}^m$, iterating $f$ will converge to a unique fixed point if $\sup_{x \in \mathbb{R}^m} \rho(f'(x)) < 1$. However, this conjecture (which has been open for a long time) has been shown to be true in two dimensions but false in higher dimensions [7].

Any matrix norm of $A$ is actually an upper bound on the spectral radius $\rho(A)$, since for any eigenvalue $\lambda$ of $A$ with eigenvector $x$ we have $|\lambda| \, \|x\| = \|\lambda x\| = \|Ax\| \leq \|A\| \, \|x\|$, hence $\rho(A) \leq \|A\|$. This implies that no norm in Lemma 3.2 will result in a sharper condition than Corollary 3.2.

## 4 General case

In the general case, when the domains $\mathcal{X}_i$ are arbitrarily large (but finite), we do not know of a natural parameterization of the messages that automatically takes care of the invariance of the messages $\mu^{I \to j}$ under scaling (like (5) does in the binary case).[4] Instead of handling the scale invariance by the parameterization and using standard norms and metrics, it seems easier to take a simple parameterization and to change the norms and metrics in such a way that they are insensitive to the (irrelevant) extra degrees of freedom arising from the scale invariance. This is actually the key insight in extending the previous results beyond the binary case: once one sees how to do this, the rest follows in a (more or less) straightforward way.

Another important point is to reparameterize the messages by switching to the logarithms of the messages $\lambda^{I \to i} := \log \mu^{I \to i}$, which results in almost linear LBP update equations, except for the "log" in front:

$$\tilde{\lambda}^{I \to i}(x_i) = \log \sum_{x_{I \setminus i}} \psi^I(x_I) h^{I \to i}(x_{I \setminus i}), \qquad (10)$$

where we dropped the normalization and defined

$$h^{I \to i}(x_{I \setminus i}) := \exp \left( \sum_{j \in I \setminus i} \sum_{J \in N_j \setminus I} \lambda^{J \to j}(x_j) \right).$$

Each log-message $\lambda^{I \to i}$ is a vector in the vector space $V_{I \to i} = \mathbb{R}^{\mathcal{X}_i}$; we will use greek letters as indices for the components, e.g. $\lambda_\alpha^{I\to i}$ with $\alpha \in \mathcal{X}_i$. We will call everything that concerns individual vector spaces $V_{I\to i}$ *local* and define the *global* vector space $V$ as the direct sum of the local vector spaces:

$$V := \bigoplus_{i\in I\in \mathcal{I}} V_{I\to i}$$

---

[4] A possible generalization, suggested by one of the reviewers, would be the parameterization in terms of $\mu^{I \to i}(x_i)/\mu^{I \to i}(x_{i_0})$ for some arbitrary $i_0$; however, this yields conditions that depend on the choice of $i_0$ and seem to be weaker in general than the results we present in this work.

The parallel LBP update is the mapping $f : V \to V$, written out in components in (10).

Note that the invariance of the messages $\mu^{I\to i}$ under scaling amounts to invariance of the log-messages $\lambda^{I\to i}$ under translation. More formally, defining linear subspaces

$$W_{I\to i} := \{\lambda \in V_{I\to i} : \lambda_\alpha = \lambda_{\alpha'} \text{ for all } \alpha, \alpha' \in \mathcal{X}_i\}$$

and their direct sum

$$W := \bigoplus_{i\in I\in \mathcal{I}} W_{I\to i} \subseteq V,$$

the invariance amounts to the observation that

$$f(\lambda + w) - f(\lambda) \in W \qquad \text{for all } \lambda \in V, w \in W.$$

Since $\lambda + w$ and $\lambda$ are equivalent for our purposes, we want our measures of distance in $V$ to reflect this equivalence. Therefore we will "divide out" the equivalence relation and work in the quotient space $V/W$, which is the topic of the next subsection.

### 4.1 Quotient spaces

Let $V$ be a finite-dimensional vector space. Let $W$ be a linear subspace of $V$. We can consider the *quotient space* $V/W := \{v+W : v \in V\}$, which is again a vector space. We will denote its elements as $\overline{v} := v + W$.

Let $\|\cdot\|$ be any vector norm on $V$. It induces a *quotient norm* on $V/W$, defined by $\|\overline{v}\| := \inf_{w\in W} \|v+w\|$, which is indeed a norm, as one easily checks. The quotient norm in turn induces the *quotient metric* $d(\overline{v_1}, \overline{v_2}) := \|\overline{v_2} - \overline{v_1}\|$ on $V/W$. The metric space $(V/W, d)$ is complete.

Let $f : V \to V$ be a (possibly non-linear) mapping with the following symmetry:

$$f(v+w) - f(v) \in W \qquad \text{for all } v \in V, w \in W. \quad (11)$$

We can then unambiguously define the quotient mapping

$$\overline{f} : V/W \to V/W : \overline{v} \mapsto \overline{f(v)}.$$

If $f$ is differentiable in $x \in V$, then the symmetry property (11) implies that $f'(x)\cdot w \in W$ for all $w \in W$.

For linear mappings $A : V \to V$ that leave $W$ invariant, i.e. $AW \subseteq W$, we can define the quotient mappings $\overline{A} : V/W \to V/W : \overline{v} \mapsto \overline{Av}$. An example is the derivative $f'(x)$ of $f$ at $x$, which induces a linear mapping $\overline{f'(x)} : V/W \to V/W$.

The operation of taking derivatives is compatible with projecting onto the quotient space. Indeed, the derivative of the induced mapping $\overline{f} : V/W \to V/W$ at $\overline{x}$ equals the induced derivative of $f$ at $x$:

$$\overline{f}'(\overline{x}) = \overline{f'(x)} \qquad \text{for all } x \in V. \quad (12)$$

By Lemma 3.2, $\overline{f}$ is a contraction with respect to the quotient norm if $\sup_{\overline{x}\in V/W} \left\|\overline{f}'(\overline{x})\right\| < 1$. We can formulate this condition more explicitly using (12):

**Lemma 4.1** *Let* $(V, \|\cdot\|)$ *be a finite-dimensional normed vector space, $W$ a linear subspace of $V$ and $f : V \to V$ a differentiable mapping satisfying (11). If*

$$\sup_{x\in V} \sup_{\substack{v\in V, \\ \|v\|\leq 1}} \inf_{w\in W} \|f'(x)\cdot v + w\| < 1, \quad (13)$$

*the induced mapping $\overline{f} : V/W \to V/W$ is a contraction with respect to the quotient norm on $V/W$.* □

### 4.2 Constructing a suitable norm

Whereas in the binary case, the local vector spaces were one-dimensional (each message $\nu^{i\to j}$ was parameterized by a single real number), the individual messages are now vectors $\lambda_\alpha^{I\to i}$ with components indexed by $\alpha \in \mathcal{X}_i$. We would like a norm that behaves "globally" like the $\ell_1$-norm (as in the binary case), but we have to generalize its local part. We give a short description of the construction; for details, see [5].

The total message space $V$ is the direct sum of linear subspaces $V_{I\to i}$. Choose a local norm $\|\cdot\|_{I\to i}$ on each subspace $V_{I\to i}$. Then we can define a global norm on $V$ simply by taking the sum of all local norms:

$$\|\lambda\| := \sum_{I\to i} \left\|\lambda^{I\to i}\right\|_{I\to i}.$$

The corresponding matrix norm for a linear $A : V \to V$ is given by

$$\|A\| = \max_{J\to j} \sum_{I\to i} \|A_{I\to i, J\to j}\|_{I\to i}^{J\to j}$$

where the norms of the matrix blocks $A_{I\to i, J\to j}$ are defined as

$$\|A_{I\to i, J\to j}\|_{I\to i}^{J\to j} := \sup_{\substack{x\in V_{J\to j}, \\ \|x\|_{J\to j}\leq 1}} \|A_{I\to i, J\to j} x\|_{I\to i}.$$

This construction is compatible with projecting onto quotient spaces; we only need to add some overlines

to get the corresponding quotient norms on $W$ and for $\overline{A}: V/W \to V/W$:

$$\|\overline{\lambda}\| = \sum_{I \to i} \left\|\overline{\lambda^{I \to i}}\right\|_{I \to i},$$

$$\|\overline{A}\| = \max_{J \to j} \sum_{I \to i} \left\|\overline{A_{I \to i, J \to j}}\right\|_{I \to i}^{J \to j} \qquad (14)$$

and

$$\left\|\overline{A_{I \to i, J \to j}}\right\|_{I \to i}^{J \to j} = \sup_{\substack{x \in V_{J \to j}, \\ \|x\|_{J \to j} \leq 1}} \left\|\overline{A_{I \to i, J \to j} x}\right\|_{I \to i}. \qquad (15)$$

### 4.3 Application to LBP updates

The global norm described in the previous section leaves open the question of how to choose the local norms. We will postpone this choice and first exploit the global properties of the norm.

The derivative of the parallel LBP update (10) is easily calculated:

$$\frac{\partial \tilde{\lambda}^{I \to i}(x_i)}{\partial \lambda^{J \to j}(y_j)} = \mathbf{1}_{N_j \setminus I}(J) \mathbf{1}_{I \setminus i}(j)$$
$$\times \frac{\sum_{x_{I \setminus i}} \psi^I(x_i, x_j, x_{I \setminus \{i,j\}}) \delta_{x_j, y_j} h^{I \to i}(x_{I \setminus i})}{\sum_{x_{I \setminus i}} \psi^I(x_i, x_{I \setminus i}) h^{I \to i}(x_{I \setminus i})}.$$

To shorten the notation, we will use greek subscripts instead of arguments: let $\alpha$ correspond to $x_i$, $\beta$ to $x_j$, $\beta'$ to $y_j$ and $\gamma$ to $x_{I \setminus \{i,j\}}$; for example, we write $h^{I \to i}(x_{I \setminus i})$ as $h_{\beta \gamma}^{I \to i}$. Taking the global quotient norm (14) of the previous expression yields:

$$\left\|\overline{f'(\lambda)}\right\| = \max_{J \to j} \sum_{I \to i} \mathbf{1}_{N_j \setminus I}(J) \mathbf{1}_{I \setminus i}(j) \times$$
$$\left\|\overline{\frac{\sum_\gamma \psi_{\alpha \beta' \gamma}^I h_{\beta' \gamma}^{I \to i}}{\sum_\beta \sum_\gamma \psi_{\alpha \beta \gamma}^I h_{\beta \gamma}^{I \to i}}}\right\|_{I \to i}^{J \to j}$$

We concentrate on the local part: fix $I \to i$ and $J \to j$ (such that $j \in I \setminus i$ and $J \in N_j \setminus I$) and drop these superscripts from the notation for the moment. Note that the local part (i.e. the fraction) depends on $\lambda$ via $h(\lambda)$. In order to get a uniform bound on $\left\|\overline{f'(\lambda)}\right\|$, we take the supremum over $h > 0$, and using (15) for the definition of the local quotient matrix norm, we arrive at the following expression:

$$\sup_{h > 0} \sup_{\substack{v \in J \to j \\ \|v\|_{J \to j} \leq 1}} \inf_{w \in W_{I \to i}} \qquad (16)$$
$$\left\|\frac{\sum_{\beta'} \sum_\gamma \psi_{\alpha \beta' \gamma} h_{\beta' \gamma} v_{\beta'}}{\sum_\beta \sum_\gamma \psi_{\alpha \beta \gamma} h_{\beta \gamma}} - w\right\|_{I \to i}.$$

Even though this expression looks rather frightening, it turns out that we can calculate it analytically if take the local norms to be $\ell_\infty$ norms. We have also tried local $\ell_2$ norms and local $\ell_1$ norms. For the $\ell_2$ norm it is possible to make some progress analytically, but we were unable to calculate the final supremum over $h$ analytically. The results of numerical calculations lead us to believe that the $\ell_2$ norm yields weaker conditions in general than the $\ell_\infty$ norm. For the $\ell_1$ norm we were only able to calculate the above expression numerically; due to the nested suprema the calculations were very time-consuming and the results were inconclusive because of many local extrema. In the next section, we present the results based on the local $\ell_\infty$ norm.

#### 4.3.1 Local $\ell_\infty$ norms

The local quotient spaces are spanned by the vector $\mathbf{1} := (1, 1, \ldots, 1)$. Taking for all local norms $\|\cdot\|_{I \to i}$ the $\ell_\infty$ norm, it is easy to see that

$$\|\overline{v}\|_{I \to i} = \frac{1}{2} \sup_{\alpha, \alpha' \in \mathcal{X}_i} |v_\alpha - v_{\alpha'}|.$$

for $v \in V_{I \to i}$.[5] It is also not difficult to see that for a linear mapping $A: V_{J \to j} \to V_{I \to i}$ that satisfies $A W_{J \to j} \subseteq W_{I \to i}$, the induced matrix quotient norm is given by

$$\left\|\overline{A}\right\|_{I \to i}^{J \to j} = \frac{1}{2} \sup_{\alpha, \alpha'} \|A_{\alpha \beta} - A_{\alpha' \beta}\|_1.$$

Hence in this case (16) equals:

$$\frac{1}{2} \sup_{\substack{h > 0, \\ \alpha, \alpha' \in \mathcal{X}_i}} \left\|\frac{\sum_\gamma \psi_{\alpha \beta \gamma} h_{\beta \gamma}}{\sum_\beta \sum_\gamma \psi_{\alpha \beta \gamma} h_{\beta \gamma}} - \frac{\sum_\gamma \psi_{\alpha' \beta \gamma} h_{\beta \gamma}}{\sum_\beta \sum_\gamma \psi_{\alpha' \beta \gamma} h_{\beta \gamma}}\right\|_1.$$

Fixing $\alpha$ and $\alpha'$, defining $\tilde{\psi}_{\beta \gamma} := \psi_{\alpha \beta \gamma}/\psi_{\alpha' \beta \gamma}$, noting that we can (without loss of generality) assume that $h$ is normalized in $\ell_1$ sense, the previous expression simplifies to

$$\frac{1}{2} \sup_{\alpha, \alpha'} \sup_{\substack{h > 0, \\ \|h\|_1 = 1}} \sum_\beta \left|\sum_\gamma h_{\beta \gamma} \left(\frac{\tilde{\psi}_{\beta \gamma}}{\sum_\beta \sum_\gamma \tilde{\psi}_{\beta \gamma} h_{\beta \gamma}} - 1\right)\right|.$$

In [5] we show that taking the supremum over $h$ yields the following analytic expression for (16):

$$N(\psi^I, i, j) :=$$
$$\sup_{\alpha \neq \alpha'} \sup_{\beta \neq \beta'} \sup_{\gamma, \gamma'} \tanh\left(\frac{1}{4} \log\left(\frac{\psi_{\alpha \beta \gamma}^I}{\psi_{\alpha' \beta \gamma}^I} \frac{\psi_{\alpha' \beta' \gamma'}^I}{\psi_{\alpha \beta' \gamma'}^I}\right)\right) \qquad (17)$$

---

[5]This expression is related to the well-known *total variation norm* in probability theory (used in [1]) and identical to the *dynamic range measure* in engineering (used in [2]).

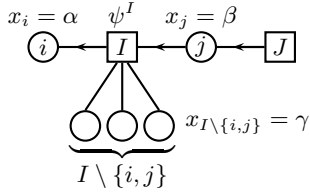

Figure 1: Part of the factor graph relevant in expressions (18) and (19). Here $i, j \in I$ with $i \neq j$, and $J \in N_j \setminus I$.

for $i, j \in I$ with $i \neq j$ and where $\psi^I_{\alpha\beta\gamma}$ is shorthand for $\psi^I(x_i = \alpha, x_j = \beta, x_{I\setminus\{i,j\}} = \gamma)$ (see also Figure 1).

Applying Lemma 4.1 yields:

**Theorem 4.1** *If*

$$\max_{J \to j} \sum_{I \in N_j \setminus J} \sum_{i \in I \setminus j} N(\psi^I, i, j) < 1, \qquad (18)$$

*LBP converges to a unique fixed point irrespective of the initial messages.* □

We can also generalize the spectral radius result Corollary 3.2. We obtain:

**Theorem 4.2** *If the spectral radius of the matrix*

$$A_{I \to i, J \to j} = \mathbf{1}_{N_j \setminus I}(J) \mathbf{1}_{I \setminus i}(j) N(\psi^I, i, j), \qquad (19)$$

*is strictly smaller than 1, LBP converges to a unique fixed point irrespective of the initial messages.*

**Proof.** Similar to the binary pairwise case. For details, see [5]. □

Note that Theorem 4.1 is a trivial consequence of Theorem 4.2. However, to prove the latter, it seems that we have to go through all the work (and some more) needed to prove the former. Also note that in case the factor graph is a tree, the spectral radius of (19) is easily shown to be zero, for any choice of the potentials, which is yet another proof of the well-known fact that LBP always converges on trees. Theorem 4.1 is not strong enough to prove that result.

We formulate our results for the important special case of pairwise interactions, which corresponds to $\gamma$ taking only one possible value. For a pair-potential $\psi^{ij}(x_i, x_j) = \psi^{ij}_{\alpha\beta}$, expression (17) simplifies to

$$N(\psi^{ij}) := \sup_{\alpha \neq \alpha'} \sup_{\beta \neq \beta'} \tanh\left(\frac{1}{4} \log\left(\frac{\psi^{ij}_{\alpha\beta} \psi^{ij}_{\alpha'\beta'}}{\psi^{ij}_{\alpha'\beta} \psi^{ij}_{\alpha\beta'}}\right)\right).$$

Note that this quantity has the desirable property that it is invariant to "reallocation" of single-node potentials $\psi^i$ or $\psi^j$ to the pair-potential $\psi^{ij}$ (i.e. $N(\psi^{ij}) = N(\psi^{ij}\psi^i\psi^j)$) and that it is symmetric in $i$ and $j$.

Condition (18) becomes for the pairwise case

$$\max_{l \in V} \max_{k \in N_l} \sum_{i \in N_l \setminus k} N(\psi^{il}) < 1, \qquad (20)$$

and the matrix (19) reduces to

$$A_{j \to i, k \to l} := N(\psi^{ij}) \delta_{j,l} \mathbf{1}_{N_j \setminus i}(k). \qquad (21)$$

For the binary case, we reobtain our earlier result Corollary 3.1, since $N\big(\exp(J_{ij} x_i x_j)\big) = \tanh |J_{ij}|$.

## 5 Discussion and conclusions

### 5.1 Comparison with other work

In [1], Tatikonda and Jordan link the convergence of LBP to the structure of the Gibbs measure on the infinite computation tree. They show that uniform convergence holds whenever the Gibbs measure is unique. To decide whether there exists a unique Gibbs measure, a classical condition from physics is invoked. In this way one obtains a practically testible sufficient condition for convergence of LBP, also known as Simon's condition.

A different approach was used in the recent work by Ihler *et. al* [2]. The authors derive bounds on the propagation of "message errors" in the network. When all message errors converge to zero, one can conclude from this that LBP converges to a unique fixed point. The authors prove that their convergence condition is stronger than Simon's condition.

In the aforementioned works, the authors limit themselves to the case of pairwise interactions, referring to a method in [4] for constructing a pairwise MRF from a Bayesian network; the resulting MRF is equivalent to the original graphical model in the sense that it has an identical probability distribution and equivalent LBP updates. However, the new MRF has deterministic potentials (i.e. potential functions containing zeroes), and applying their sufficient conditions to the pairwise MRF, it seems that one obtains conditions that are never satisfied.

The condition in [2] for convergence of LBP is almost the same as (20), except for the quantity $N(\psi^{ij})$, which is replaced by:

$$D(\psi^{ij}) := \tanh\left(\frac{1}{2}\left(\sup_{\alpha,\beta} \log \psi^{ij}_{\alpha\beta} - \inf_{\alpha',\beta'} \log \psi^{ij}_{\alpha'\beta'}\right)\right).$$

It is not difficult to show that $N(\psi^{ij}) \leq D(\psi^{ij})$ for any pair-potential $\psi^{ij}$, hence our condition (20) is sharper than that in [2]. In general the difference of these

quantities can be quite large.[6]

Finally we would like to mention the work of Heskes [3] in which sufficient conditions on the *uniqueness* of the LBP fixed point are derived by a careful analysis of the properties of the Bethe free energy. The conjecture is made that uniqueness of the fixed point implies convergence of LBP. It is not known how these conditions exactly relate to the ones presented in this work and this may be an interesting question for further research.

### 5.2 Comparison with empirical LBP convergence results

In Figure 2 we compare our sufficient conditions with the empirical convergence behaviour of parallel, undamped LBP. The graphical model is a rectangular 2D grid of size $10 \times 10$ with periodic boundary conditions, binary variables, random independent and normally distributed nearest-neighbour interactions $J_{ij} \sim \mathcal{N}(J_0, J)$ and no local evidence (i.e. $\theta_i = 0$). For several different ratios of $J_0/J$, the results have been averaged over 40 instances of the network; the lines correspond to the means, the gray areas are "error bars" of one standard deviation. Note that the spectral radius condition appears to be sharp when most of the weights $J_{ij}$ have similar sign.

### 5.3 Open questions

We have assumed from the outset that all potentials are positive. However, one may observe that the spectral radius condition also makes sense for deterministic (non-negative) potentials (using that $\lim_{x \to \infty} \tanh x = 1$). This suggests that the result also holds generally for non-negative potentials, although we currently have no prove of this conjecture.

Other questions raised by this work are how the conditions change for damped LBP or alternative (e.q. sequential) update schemes and how the analysis could be extended to Generalized Belief Propagation or other message passing algorithms.

---

[6]After submission of this work, we became aware of a recent extension of the results of Ihler *et al.*, given in [8]. By exploiting the freedom in the choice of the single-node and the pairwise potentials, they improved their bounds and in this way obtained a result identical to (21), for the pairwise case. Also, as observed by Ihler [personal communication], the improved version of the non-uniform ("path-counting") distance bound [8, Thm. 5.5] is equivalent to our spectral radius bound, again for the pairwise case. This can be proved using Theorem 3.2.

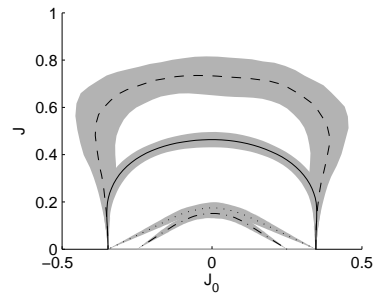

Figure 2: Comparison of the empirical convergence behaviour of LBP (dashed line), the spectral radius condition Corollary 3.2 (solid line), the $\ell_1$-norm based condition Corollary 3.1 (dotted line) and Simon's condition (dash-dotted line). The inner areas (i.e. around the origin) mean "convergence" (either guaranteed or empirical). See the main text for more explanation.

### Acknowledgements

The research reported here is part of the Interactive Collaborative Information Systems (ICIS) project, supported by the Dutch Ministry of Economic Affairs, grant BSIK03024. We thank the anonymous reviewers, Martijn Leisink and Tom Heskes for helpful discussions and comments.